\documentclass[twoside]{article}
\usepackage[accepted]{aistats2015}
\usepackage{subcaption}
\usepackage{bbm}

\newcommand{\ignore}[1]{}
\usepackage{algorithm}
\usepackage{algorithmic}
\usepackage{amsmath}
\usepackage{amsfonts}
\usepackage{natbib}
\usepackage{graphicx} 
\usepackage{enumitem}
\usepackage{mathtools}
\usepackage{enumitem}
\usepackage{hyperref}
\usepackage{xfrac}
\setlist{nolistsep}
\begin{document}

\twocolumn[

\aistatstitle{Falling Rule Lists}

\aistatsauthor{ Fulton Wang \And Cynthia Rudin }

\aistatsaddress{CSAIL and Sloan, MIT \\ \texttt{fultonw@mit.edu} \And CSAIL and Sloan, MIT\\ \texttt{rudin@mit.edu}}
]
\begin{abstract}
Falling rule lists are classification models consisting of an ordered list of if-then rules, where (i) the order of rules determines which example should be classified by each rule, and (ii) the estimated probability of success decreases monotonically down the list. These kinds of rule lists are inspired by healthcare applications where patients would be stratified into risk sets and the highest at-risk patients should be considered first. We provide a Bayesian framework for learning falling rule lists that does not rely on traditional greedy decision tree learning methods.
\end{abstract}

%\end{frontmatter}

\section{Introduction}
In healthcare, patients and actions need to be prioritized based on risk. The most at-risk patients should be handled with the highest priority, patients in the second at-risk set should receive the second highest priority, and so on. This decision process is perfectly natural for a human decision-maker -- for instance a physician -- who might check the patient for symptoms of high severity diseases first, then check for symptoms of less serious diseases, etc.; however, the traditional paradigm of predictive modeling does not naturally contain this type of logic. If such clear logic were well-founded, a typical machine learning model would not usually be able to discover it: most machine learning methods produce highly complex models, and were not designed to provide an ability to reason about each prediction. This leaves a gap, where predictive models are not directly aligned with the decisions that need to be made from them. 

The algorithm introduced in this work 
%\textit{falling rule lists}, 
aims to resolve this problem, and could be directly useful for clinical practice. A \textit{falling rule list} is an ordered list of if-then rules, where (i) the order of rules determines which example should be classified by each rule (falling rule lists are a type of decision list), and (ii) the estimated probability of success decreases monotonically down the list. Thus, a falling rule list directly contains the decision-making process, whereby the most at-risk observations are classified first, then the second set, and so on. A falling rule list might say, for instance, that patients with a history of heart disease are in the highest risk set with a  7\% stroke risk, patients with high blood pressure (who are not in the highest risk set) are in the second highest risk set with a 4\% stroke risk, and patients with neither conditions of these are in the lowest risk set with a 1\% stroke risk. 

Table \ref{fig:mammolist} shows an example of one of the decision lists we constructed for the mammographic mass dataset \citep{elter2007prediction} as part of our experimental study. It took 35 seconds to construct this model on a laptop. The model states that if biopsy results show that the tumor has irregular shape, and the patient is over age 60, then the tumor is at the highest risk of being malignant (the risk is 85\%). The next risk set is for the remaining tumors that have spiculated margins and are from patients above 45 years of age (the risk is 78\%), and so on. The right column of Table \ref{fig:mammolist} shows how many patients fit into each of the rules (so that its sum is the size of the dataset), and the risk probabilities were directly calibrated to the data.
\begin{table*}[]
\begin{tabular}{lllll}
       & Conditions                         &                        & Probability & Support \\\hline
IF      & IrregularShape AND Age $\geq$ 60   & THEN malignancy risk is & 85.22\% & 230     \\
ELSE IF & SpiculatedMargin AND Age $\geq$ 45  & THEN malignancy risk is & 78.13\%     & 64      \\
ELSE IF & IllDefinedMargin AND Age $\geq$ 60 & THEN malignancy risk is & 69.23\% & 39      \\
ELSE IF & IrregularShape                     & THEN malignancy risk is & 63.40\% & 153     \\
ELSE IF & LobularShape AND Density $\geq$ 2  & THEN malignancy risk is & 39.68\% & 63      \\
ELSE IF & RoundShape AND Age $\geq$ 60        & THEN malignancy risk is & 26.09\% & 46      \\
ELSE    &                                   & THEN malignancy risk is & 10.38\% & 366    
\end{tabular}
\caption{Falling rule list for mammographic mass dataset.\label{fig:mammolist}}
\end{table*}

%A falling rule list might say, for instance, that patients with a history of heart disease might be in the highest risk category with a  7\% stroke risk, patients with high blood pressure are in the second highest risk category with a 4\% stroke risk, and patients with neither of these are in the lowest risk category with a 1\% stroke risk. 

Falling rule lists serve a dual purpose: they rank rules to form a predictive model, and stratify patients into decreasing risk sets. This saves work for a physician; sorting is an expensive mental operation, and this model does the sorting naturally. If one were to use a standard decision tree or decision list method instead, identifying the highest at-risk patients could be a much more involved calculation, and the number of conditions the most at-risk patients need to satisfy might be difficult for physicians to memorize.
% Note that one could argue with this. If we used a pruned decision tree, then we could pluck out the highest at-risk patient categories, and it wouldn't be much to memorize since the whole tree is shallow.

Most of the models currently in use for medical decision making were designed by medical experts rather than by data-driven or algorithmic approaches. These manually-created risk assessment tools are used in possibly every hospital; e.g., the TIMI scores, CHADS$_2$ score, Apache scores, and the Ranson score, to name a few \citep{antman2000timi, morrow2000timi,gage2001validation,knaus1981apache,knaus1985apache,knaus1991apache,ranson1974prognostic}.
% criminology instruments that are widely used to assess the risk of violence \cite{andrade2009handbook,steinhart2006juvenile}, tools that gauge marine safety for military vessels \cite{abs2002marine}, and business school ratings  \cite{business}. 
These models can be computed without a calculator, making them very practical as decision aids. Of course, we aim for this level of interpretability in purely data-driven classifiers, with no manual feature selection or rounding coefficients.

%Algorithms that discretize the input space have gained in popularity purely because they yield interpretable models. Inductive logic programming \citep{muggleton1994inductive, muggleton1995inverse, muggleton2010progolem} aims to find a disjunctive hypothesis to explain a collection of positive examples, returning an unstructured set of conjunctive rules such that an example is classified as positive if it satisfies any of the rules in that set.  One drawback of ILP methods is the lack of an underlying probabilistic model.  Recent approaches \citep{landwehr2005nfoil, de2008probabilistic} have addressed this by, for example, incorporating ILP into a naive Bayes framework.  Alternatively, an extremely simple way to induce a probabilistic model from the unordered set of rules given by an ILP method is to place them into a decision list \citep{rivest1987learning,fawcett2008prie}, ordering the rules by their respective empirical risks.  However, the resulting model cannot be expected to exhibit good predictive performance, as its constituent rules were chosen with a different objective in mind.  Alternatively, decision trees \citep{Breiman84,quinlan1986induction,quinlan1993c4}, as well as decision lists, directly optimize the likelihood of models that organize a collection of simple rules into a larger logical structure, and are popular despite performing the optimization in a greedy manner.

Algorithms that discretize the input space have gained in popularity purely because they yield interpretable models. Decision trees \citep{Breiman84,quinlan1986induction,quinlan1993c4}, as well as decision lists \citep{rivest1987learning}, organize a collection of simple rules into a larger logical structure, and are popular despite being greedy. Inductive logic programming \citep{muggleton1994inductive} returns an unstructured set of conjunctive rules such that an example is classified as positive if it satisfies any of the rules in that set. An extremely simple way to induce a probabilistic model from the unordered set of rules given by an ILP method is to place them into a decision list \citep[e.g., see][]{fawcett2008prie}, ordering rules by empirical risk. This is also done in associative classification \citep[e.g., see][]{Thabtah07}. However, the resulting model cannot be expected to exhibit good predictive performance, as its constituent rules were chosen with a different objective.

%.  Directly optimizing the likelihood by organizing simple rules in a decision list or decision tree structure is a popular approach, despite being greedy, possibly due to their interpretability.

%Decision tree \citep{Breiman84,quinlan1986induction,quinlan1993c4} and decision list \citep{rivest1987learning,fawcett2008prie}  methods organize a set of typically simple rules into a larger logical structure, and are popular despite being greedy and not as accurate as other algorithms.
Since it is possible that decision tree methods can produce results that are inconsistent with monotonicity properties of the data, there is a subfield dedicated to altering these greedy decision tree algorithms to obey monotonicity properties \citep{ben1995monotonicity,feelders2003pruning,altendorf2012learning}. Studies showed that in many cases, no accuracy is lost in enforcing monotonicity constraints, and that medical experts were more willing to use the models with the monotonicity constraints \citep{pazzani2001acceptance}. 

Even with (what seem like) rather severe constraints on the hypothesis space such as monotonicity or sparsity in the number of leaves and nodes, it still seems that the set of accurate classifiers is often large enough so that it contains interpretable classifiers \citep[see][]{holte1993very}. Because the monotonicity properties we enforce are much stronger than those of \citet{ben1995monotonicity,feelders2003pruning,altendorf2012learning} (we are looking at monotonicity along the whole list rather than for individual features), we do find that accuracy is sometimes sacrificed, but not always, and generally not by much. On the other hand, it is possible that our method gains a level of practicality and interpretability that other methods simply cannot.

Interpretability is very context dependent
\citep[see][]{kodratoff1994comprehensibility,pazzani2000knowledge, Freitas:2014ic,huysmans2011empirical,allahyari2011user, martens2010building, ruping2006learning,verbeke2011building,martens2011performance}, and no matter how one measures it in one domain, it can be different in the next domain. A falling rule list used in medical practice has the benefit that it can, in practice, be as sparse as desired. Since it automatically stratifies patients by risk in the order used for decision making, physicians can choose to look at as much of the list as they need to make a decision; the list is as sparse as one requires it to be.  
If physicians only care about the most high risk patients, they look only at the top few rules, and check whether the patient obeys any of the top clauses. 
%Depending on the situation, the physician might look only at the top rule. 
%To classify the most at-risk patients, physicians need only to keep small amounts of information memorized (namely the first few rules). That is, the amount of information a physician needs to memorize and check 

%They see a set of clauses, and say ok if someone has one of these, then they're high risk.  If they wanted to they could just look at the very top rule.  One has to keep very few things in their head at once.  The number of rules one has to look at decreases with the number of patients the doctor wishes to identify as being high risk.
%On the other hand, with a traditional decision list might have a high risk probability near the bottom, in which case a doctor has to look at the clause for the node with a high risk probability, and mentally subtract out *all* of the clauses above it.  That's potentially a lot of things to keep in one's head, just to figure out what kind of patient is most at risk.
%Maybe we could emphasize the number of things a person has to keep in their head to understand the relationship between risk and patient features.
%In our model, monotonicity is enforced fully enforced , and the desired level of sparsity for the full list is captured in a soft way through a Bayesian prior. 

The algorithm we provide for falling rule lists aims to have the best of all worlds: accuracy, interpretability, and computation. The algorithm starts with a statistical assumption, which is that we can build an accurate model from pre-mined itemsets. This helps tremendously with computation, and restricts us to building models with only interpretable building blocks \citep[see also][]{LethamRuMcMa14,WangRuDoLiKlMa14}. Once the itemsets are discovered, a Bayesian modeling approach chooses a subset and permutation of the rules to form the decision list. The user determines the desired size of the rule list through a Bayesian prior. Our generative model is constructed so that the monotonicity property is fully enforced (no ``soft" monotonicity).

The code for fitting falling rule lists is available online\footnote{\url{http://web.mit.edu/rudin/www/falling_rule_list}}.

  %\textbf{We have made the Python code for fitting a falling rule list available online \footnote{http://web.mit.edu/rudin/www/falling_rule_list}}. 
% and not able to be overwhelmed by data.

\section{Falling Rule Lists Model}
We consider binary classification, where the goal is to learn a
distribution $p(Y|x)$, where $Y$ is binary. For example, $Y$ might indicate the presence of a disease, and $x$ would be a patient's features. We represent this
conditional distribution as a decision list, which is an ordered list of IF...THEN... rules. We require a special structure to this decision list: that the probability of $Y=1$ associated with each rule is \emph{decreasing} as one moves down the decision list. 

We use a Bayesian approach to characterize the posterior over falling rule lists given training data $D=\{(x_n, y_n)\}_{n=1,\dots,N}$  (of size $N$), $x_n \in X$, the patient feature space, hyperparameters $H$,
and $y_n \in \{0,1\}$.   We represent a falling rule list with a set of parameters $\theta$, specify the prior $p_{\theta}(\cdot;H)$ and likelihood $p_{\mathbf{Y}}(\{y_n\}|\theta;\{x_n\} )$, and use simulated annealing and Monte Carlo sampling to approximate the MAP estimate and posterior over falling rule lists,% $p_{\theta}(\theta|\{y_n\};\{x_n\},H)$. 

\subsection{Parameters of Model}
%We first introduce the parameterization of a falling decision list, the restrictions on the parameters, and the likelihood function.  We then describe a prior on those parameters respecting the restrictions placed on them.

%\subsection{Decision list parameterization}

A falling rule list is parameterized by the following objects:
%\begin{flalign}
%L &\in \mathbb{Z^+} &\textrm{(size of list)}&\\
%c_l(\cdot) \in B_X(\cdot), &\text{ for} \quad l = 0, \dots, L-1 & \textrm{(IF clauses)}&\\
%\{c_l&(\cdot)\}_{l=0}^{L-1} \} &\textrm{where} \nonumber&\\
%c_l(\cdot) &\in B_X(\cdot), \text{ for} \quad l = 0, \dots, L-1 & \textrm{(IF clauses)}&\\
%r_l &\in \mathbb{R}, \text{ for} \quad l = 0, \dots, L& \textrm{(risk scores)}&\\
%\intertext{such that}
%r_{l+1} &\leq r_l \text{ for} \quad  l=0,\dots,L-1& \textrm{(monotonic)}
%\end{flalign}
\begin{flalign}
L &\in \mathbb{Z^+}&&\text{(size of list)}\\
c_l(\cdot) &\in B_X(\cdot), \text{ for} \quad l = 0, \dots, L-1 && \text{(IF clauses)}\\
r_l &\in \mathbb{R}, \text{ for} \quad l = 0, \dots, L &&\text{(risk scores)}\\
\shortintertext{such that}
r_{l+1} &\leq r_l \text{ for} \quad  l=0,\dots,L-1 &&\text{(monotonic)}
\end{flalign}
% \vspace*{0pt}
where $B_X(\cdot)$ denotes the space of boolean functions on patient feature space $X$. $B_X(\cdot)$ is the space of possible IF clauses; $c_l(x)$ will be 1 if $x$ satisfies a given set of conditions.
For this work, we will not assume $L$, the size of the decision list, to be known in advance. The value of $r_l$ will be fed into the logistic function to produce a risk probability between 0 and 1.  Thus, $c_0(\cdot)$ corresponds to the rule at the top of the list, determining the patients with the highest risk, and there are $L+1$ nodes and associated risk probabilities in the list: $L$ associated with the $c_l(\cdot)$'s, plus one for \emph{default} patients - those matching none of the $L$ rules.
% positive if the predicted risk for $Y=1$ to be above probability 1/2. 

\subsection{Likelihood}

Given these parameters, the likelihood is as follows: Given $L$, let $Z(x;\{c_l(\cdot)\}_{l=0}^{L-1}): X \to \{0,\dots,L\}$ be the mapping from feature $x$ to the index of the length $L$ rule list it ``belongs'' to (equals $L$ for default patients):
\begin{align}
\lefteqn{Z(x;\{c_l(\cdot)\}_{l=0}^{L-1}) = }\\ 
&\begin{cases}
L  \text{\;\;\;\;\;\;\;\;\;\;\;\;\;\;\;if} \enskip c_l(x)=0 \enskip \text{for} \enskip l=0,\dots,L-1 \nonumber\\
\min(l: c_l(x)=1, \enskip l=0,\dots,L-1)  \text{\;\;otherwise}.
\end{cases}
\end{align}
Then, the likelihood is:
\begin{eqnarray}
\lefteqn{y_n|L, \{c_l(\cdot)\}_{l=0}^{L-1},\{r_l\}_{l=0}^L; x_n \sim} \nonumber\\ &\operatorname{Bernoulli}(\operatorname{logistic}(r_{z_n})), \enskip \text{where}\\
&z_n = Z(x_n;\{c_l(\cdot)\}_{l=0}^{L-1}). \label{eqn:z_n}
\end{eqnarray}

%Then, the likelihood is:
%\begin{align}
%y_n|\Theta,; x_n &\sim \operatorname{Bernoulli}(\operatorname{logistic}(r_l)), &\textrm{ where}\\
%\intertext{where}
%p_l &= \operatorname{logistic}(r_l), &\textrm{and} \\
%z_n &= Z(x_n;\{c_l(\cdot)\}_{l=0}^{L-1}). \label{eqn:z_n}
%\end{align}

\subsection{Prior}

Here, we describe the prior over the parameters $L$, $\{c_l\}_{l=0}^{L-1}, \{r_l\}_{l=0}^L$.
%, which uses the domain of boolean functions $B:X\rightarrow \{0,1\}$ that we permit in the list. 
We will provide a reparameterization that enforces monotonicity constraints, and finally give a generative model for the parameters.

As discussed earlier, to help with computation, we place positive prior probability of $\{c_l\}_{l=0}^{L-1}$ only over lists consisting of boolean clauses $B$ returned by a frequent itemset mining algorithm, where for $c(\cdot)\in B$, we have $c(\cdot):X\rightarrow \{0,1\}$.  
%(Note that Bayesian decision tree methods, e.g., \citealt{Chipman:1998jh}, make a different assumption, which is to assign positive probability only to trees that can be constructed greedily).  
For this particular work we used FPGrowth \citep{borgelt2005implementation}, whose input is a binary dataset where each $x$ is a boolean vector, and whose output is a set of subsets of the features of the dataset.  For example, $x_2$ might indicate the presence of diabetes, and $x_{15}$ might indicate the presence of hypertension, and a boolean function returned by FPGrowth might return $1$ for patients who have diabetes and hypertension.  It does not matter which rule mining algorithm is chosen because they all perform breadth-first searches to return a set of clauses that have sufficient support.  Here, $B$ needs to be sufficiently large, so that the hypothesis space of considered models is not too small.  $B$ can be viewed as a hyperparameter, and the maximum length a decision list can have under our model is $|B|$, the number of rules in $B$.%Also, note that the maximum length a decision list can have under our model is $|B|$, the number of clauses returned by the rule mining algorithm, and that $B$ can be viewed as a hyperparameter.

\subsubsection{Reparameterization}
To ensure the monotonicity constraints that $r_l\geq
r_{l-1}$ for $l=1\dots L$ in the posterior, we choose the scores $r_l$ to, on a log scale, come from products of real numbers constrained to be greater than 1. %Thus, $r_{l-1}$ is larger than $r_{l}$ by a factor that is larger than 1. 
That is, conditioned on
$L$, we let 
\begin{align}
r_l &= \log(v_l) \enskip &\text{for}& \enskip l=0,\dots,L \\
v_l &= K\Pi_{\grave{l}=l}^{L-1} \gamma_{\grave{l}} \enskip &\text{for}& \enskip l=0,\dots,L-1 \\
v_L &= K \\
\intertext{and require that}
\gamma_l &\geq 1 \enskip &\text{for}& \enskip l=0,\dots,L-1 \label{eqn:monotonic}\\
K &\geq 0,
\end{align}
so that $r_L$, the risk score associated with the default rule, equals $\log K$.  The prior we place over $\{\gamma_l\}_{l=0}^{L-1}$ and $K$ will respect those constraints.

Thus, after reparameterizing, the parameters are 
\begin{eqnarray}
\theta = \{L, \{c_l(\cdot)\}_{l=0}^{L-1}, \{\gamma_l\}_{l=0}^{L-1}, K\} \label{eqn:params}.
\end{eqnarray}
\vspace*{-15pt}
%It is easily checked that these constraints on $\gamma_l$ ensure the
%desired monotonicity conditions in $r_l$.

\subsubsection{Prior Specifics}
The prior over parameters $L$, $\{c_l(\cdot)\}_{l=0}^{L-1}, \{\gamma_l\}_{l=0}^{L-1}, K$ is generated through the following process:\vspace*{1mm}\\
%\begin{enumerate}
% \item 
1.  Let hyperparameters \\$H = \{B, \lambda, \{\alpha_l\}_{l=0}^{|B|-1}, \{\beta_l\}_{l=0}^{|B|-1}, \alpha_K, \beta_K, \{w_l\}_{l=0}^{|B|-1}\}$ be given.\\
2. Initialize $\Theta \leftarrow \{\}.$\\
3. Draw $L \sim \operatorname{Poisson}(\lambda).$\\
4. For $l=0,\dots,L-1$ draw \vspace*{-0mm}\\
%   where
\vspace*{-5mm}
   \begin{eqnarray}
   \lefteqn{c_l(\cdot) \sim p_{c(\cdot)}\left(\cdot|\Theta;B,\{w_l\}_{l=0}^{|B|-1}\right)} \vspace*{-5mm}\\
    \lefteqn{ p_{c(\cdot)}\;\left(c(\cdot)=c_j(\cdot)|\Theta;B,\{w_l\}_{l=0}^{|B|-1}\right)} \vspace*{-5mm}\\
    & \propto w_j \textrm{ if } c_j(\cdot) \notin \Theta \textrm{ and } 0 \textrm{ otherwise.} \vspace*{-5mm}\label{eqn:pc}\\
    \lefteqn{\text{Update} \enskip \Theta \leftarrow \Theta \cup \{c_l(\cdot).\}}
   \end{eqnarray} \vspace*{-6mm}\\
5. For $l=0,\dots,L-1$ draw $\gamma_l \sim \operatorname{Gamma_1}(\alpha_l,\beta_l)$, which is a Gamma distribution truncated to have support only above 1.\vspace*{0mm}\\
6. Draw $K \sim \operatorname{Gamma}(\alpha_K, \beta_K)$.\vspace*{-3mm}\\
%\end{enumerate}

% \begin{enumerate}[topsep=0pt,itemsep=-3.5ex,partopsep=1ex,parsep=1ex]
% %\begin{enumerate}[nolistsep]
% \item Let hyperparameters $H = \{B, \lambda, \{\alpha_l\}_{l=0}^{|B|-1}, \{\beta_l\}_{l=0}^{|B|-1}, \alpha_K, \beta_K, \{w_l\}_{l=0}^{|B|-1}\}$ be given.\\
% \item Initialize $\Theta \leftarrow \{\}.$\\
% \item Draw $L \sim \operatorname{Poisson}(\lambda).$\\
% \item For $l=0,\dots,L-1$ 
%   \begin{itemize}
%     \item[] draw $c_l(\cdot) \sim p_{c(\cdot)}(\cdot|\Theta;B,\{w_l\}_{l=0}^{|B|-1})$ where \\
% \begin{eqnarray}
%      p_{c(\cdot)}\;(c(\cdot)=c_j(\cdot)|\Theta;B,\{w_l\}_{l=0}^{|B|-1})  \propto 
% \begin{cases} 
% w_j \textrm{ if } c_j(\cdot) \in B \setminus \Theta\\
% 0 \enskip \textrm{otherwise.} \label{eqn:pc}
% \end{cases}
% \end{eqnarray}
%    \item[] Update $\Theta \leftarrow \Theta \cup c_l(\cdot)$.\\
%      \end{itemize}
% \item For $l=0,\dots,L-1$ 
% \begin{itemize}
% \item[] draw $\gamma_l \sim \operatorname{Gamma_1}(\alpha_l,\beta_l)$ 
% \end{itemize}
% where $\operatorname{Gamma_1}$ is a Gamma distribution truncated to have support only above 1.\\
% \item Draw $K \sim \operatorname{Gamma}(\alpha_K, \beta_K)$.\\
% \end{enumerate}

We now elaborate on our choice for each involved distribution.  We let $L \sim \operatorname{Poisson}(\lambda)$, where $\lambda$ reflects the prior decision length desired by the user.  We let $c_l(\cdot)$ be the result of a draw from a discrete distribution over the yet unchosen rules, $B \setminus \{c_{\grave{l}}(\cdot)\}_{\grave{l}=0}^{l-1}$, where the $l$-th rule is drawn with probability proportional to a user designed weight $w_l$.  For example, a rule might be chosen with probability proportional to the number of clauses in it. This allows the user to express preferences over different types of clauses in the list. Given $L$, only $\{c(\cdot)_l\}_{l=1}^{L-1}$ are observed, though note this process specifies a joint distribution over all of $\{c(\cdot)_l\}_{l=1}^{|B|-1}$.  Letting $\{\gamma_l\}_{l=0}^{L-1}$ to be independently distributed truncated gamma variables permits posterior Gibbs sampling while enforcing the monotonicity constraints and still permitting diversity over prior distributions.  For example, one could encourage some of the $\gamma$'s near the middle (of $L$) of the list to be large, in which case the risks would be widely spaced in the middle of the list (but this would force closer spacing at the top of the list where the risks concentrate near 1).  Finally, $K$, which models the risk of patients not satisfying any rules, is Gamma distributed.% in the prior.%from an (untruncated) Gamma distribution in the prior to %facilitate posterior sampling.

\section{Fitting the Model}

First we describe our approach to finding the decision list with the maximum a posteriori probability. Then we discuss our approach to perform Monte Carlo sampling from the posterior distribution over decision list parameters $\theta=\{L,c_{0,\dots,L-1}(\cdot),K,\gamma_{0,\dots,L-1}\}$ as described in Equation (\ref{eqn:params}), 
\begin{equation}
p_{\operatorname{post}}(L,c_{0,\dots,L-1}(\cdot),K,\gamma_{0,\dots,L-1}|y_{1,\dots,N};\mathbf{x}_{1,\dots,N}). \label{eqn:posterior}
\end{equation}

\subsection{Obtaining the MAP decision list \label{sec:map}}

We adopted a simulated annealing approach to find
 $\theta^*=\{L^*,c_{0,\dots,L^*-1}(\cdot)^*,K^*,\gamma_{0,\dots,L^*-1}\}$, where
\begin{eqnarray}
\lefteqn{L^*, c^*_{0,\dots,L^*-1}(\cdot), K^*, \gamma^*_{0,\dots,L^*-1}}\label{eqn:max_eqn}\\
&\in \operatorname{argmax}_{L,c_{0,\dots,L-1}(\cdot),K,\gamma_{0,\dots,L-1}} \mathcal{L} \nonumber
\end{eqnarray}
where $\mathcal{L}$ is shorthand for the unnormalized log of the posterior given in Equation (\ref{eqn:posterior}).
We note that the optimization problem in Equation (\ref{eqn:max_eqn}) is equivalent to finding:
\begin{align}
\lefteqn{L^*,c_{0,\dots,L^*-1}(\cdot)^*}\label{eqn:alt_max_eqn}\\
&\in \operatorname{argmax}_{L,\{c_l(\cdot)\}_{l=0}^{L-1}} \mathcal{L}(L, \{c_l(\cdot)\}_{l=0}^{L-1}, K^*, \gamma_{0,\dots,L-1}^*) \nonumber
\end{align}
where
\begin{align}
\lefteqn{K^*,\gamma_{0,\dots,L-1}^*}\\
%\\&%\hspace*{-10pt}
&\in \operatorname{argmax}_{K,\gamma_{0,\dots,L-1}}\mathcal{L}(L, \{c_l(\cdot)\}_{l=0}^{L-1}, K, \gamma_{0,\dots,L-1}).\nonumber
\end{align}
Note that $K^*$ and $\gamma_{0,\dots,L-1}^*$ depend on $L, \{c_l(\cdot)\}_{l=0}^{L-1}$.

Furthermore, the solution to the subproblem of finding $K^*$ and $\gamma_{0,\dots,L-1}^*$ can be approximated closely using a simple procedure, as it involves maximizing the posterior probability of a decision list given the rules $\{c_l(\cdot)\}_{l=0}^{L-1}$.  Optimizing Equation (\ref{eqn:alt_max_eqn}) lends itself better to simulated annealing than Equation (\ref{eqn:max_eqn}); optimizing (\ref{eqn:alt_max_eqn}) involves optimization over a discrete space, namely the set and order of rules $c_l(\cdot)$. In this formulation, at each simulated annealing iteration, we need to evaluate the objective function for the current rule list $\{c_l(\cdot)\}_{l=0}^{L-1}$, which involves solving the continuous subproblem of finding the corresponding $K^*$ and $\gamma_{0,\dots,L-1}^*$.  

Given a objective function $E(s)$ over discrete search space $S$, a function specifying the set of neighbors of a state $N(s)$, and a temperature schedule function over time steps, $T(t)$, a simulated annealing procedure is a discrete time, discrete state Markov Chain $\{s_t\}$ where at time $t$, given the current state $s_t$, the next state $s_{t+1}$ is chosen by first randomly selecting a proposal $\grave{s}$ from the set $N(s)$, and setting $s_{t+1}=\grave{s}$ with probability $\min(1, \exp(-\frac{E(\grave{s}) - E(s)}{T(t)}))$, and $s_{t+1}=s_t$ otherwise.

The search space of the optimization problem of Equation (\ref{eqn:alt_max_eqn}) is $\left\{L,\{c_l(\cdot)\}_{l=0}^{L-1}\right\}$, the set of ordered lists of rules drawing from the finite pre-mined set of rules $B$.  Based on Equation (\ref{eqn:alt_max_eqn}), we let 
\begin{flalign}
\hspace*{-2mm}  E(\{c_l(\cdot)\}_{l=0}^{L-1}) = -\mathcal{L}(L, \{c_l(\cdot)\}_{l=0}^{L-1}, K^*, \gamma_{0,\dots,L-1}^*). \hspace*{-2mm}
\end{flalign}
We simultaneously define the set of neighbors and the process by which to randomly choose a neighbor through the following random procedure that alters the current rule list $\{c_l(\cdot)\}_{l=0}^{L-1}$ to produce a new rule list $\{\grave{c}_l(\cdot)\}_{l=0}^{\grave{L}-1}$ (The new list's length may be different):

\itemsep0em

Choose uniformly at random one of the following 4 operations to apply to the current rule list, $\{c_l(\cdot)\}_{l=0}^{L-1}$:
\vspace{-8mm}
\begin{enumerate}[topsep=10pt,itemsep=0.5ex,partopsep=11ex,parsep=1ex]
\item SWAP: Select $i \neq j$ uniformly from $0,\dots,L-1$, and swap the rules at those 2 positions, letting $\grave{c_i}(\cdot) \leftarrow c_j(\cdot)$ and $\grave{c_j}(\cdot) \leftarrow c_i(\cdot)$. \vspace*{-5mm}\\
%  \item 

\item REPLACE: Select $i$ uniformly from $0,\dots,L-1$, draw $c(\cdot)$ from the the distribution\\ $p_{c(\cdot)}(\cdot|\Theta;B,\{w_l\}_{l=0}^{|B|-1})$ defined in Equation (\ref{eqn:pc}), where $\Theta=\{c_l(\cdot)\}_{l=0,\dots,i-1,i+1,\dots,L-1}$ and set $\grave{c}_i(\cdot) \leftarrow c(\cdot)$.\vspace*{-5mm}\\
\item ADD: Choose one of the $L+1$ possible insertion points uniformly at random, draw
a rule $c(\cdot)$ from $p_{c(\cdot)}(\cdot|\Theta;B,\{w_l\}_{l=0}^{|B|-1})$, where $\Theta=\{c_l(\cdot)\}_{l=0,\dots,L-1}$, and insert it at the chosen insertion point, so that $\grave{L} \leftarrow L+1$.  \vspace*{-5mm}\\
\item REMOVE: Choose $i$ uniformly at random from $0,\dots,L-1$, and remove $c_i(\cdot)$ from the current rule list, so that $\grave{L} \leftarrow L-1$.
\end{enumerate}\vspace*{-5mm}

Note that this approach optimizes over the full set of rule lists from itemsets, and does not rely on greedy splitting. Even Bayesian tree methods that aim to traverse a wider search space use greedy splitting and local solutions, e.g. \citet{Chipman:1998jh}.

\subsection{Obtaining the posterior}

To perform posterior sampling, we use Gibbs sampling steps over $\{\gamma_l\}_{l=0}^{L-1}$ and $K$
made possible by variable augmentation, and Metropolis-Hastings steps
over $L$ and $\{c_l(\cdot)\}$.  We describe the variable augmentation step, the schedule of updates we employ, and finally the details of each individual update step.

Augmenting the model
with two additional variables $U_n, \zeta_n$ for each $n=1,\dots,N$
preserves the marginal distribution over the original variables, and enables Gibbs
sampling over $K$ and each $\gamma_l$ \citep[see][]{dunson2004bayesian}:
\begin{align}
\zeta_n & \sim \operatorname{Exponential}(1) \enskip &\text{for}&
\enskip n=1,\dots,N \\
U_n & \sim \operatorname{Poisson}(\zeta_n v_{z_n}) \enskip &\text{for}& \enskip
n=1,\dots,N \\
Y_n & = 1(U_n>0) \enskip &\text{for}&
\enskip n=1,\dots,N. \label{eqn:w}
\end{align}
Marginalizing over $\zeta_n$, we see that in this augmented
model, $y_n \sim
\operatorname{Bernoulli}(\operatorname{logistic}(r_{z_n}))$, as before:
\begin{align}
p(Y_n=1) &= p(U_n > 0) \\
& = 1 - \int_0^{\infty} p(U_n = 0 | \zeta_n) p(\zeta_n) d \zeta_n \\
& = 1 - \int_0^{\infty} \exp(-\zeta_n v_{z_n}) \exp(-\zeta_n) d \zeta_n \\
& = 1 - (1+v_{z_n})^{-1} \\
%& = \frac{v_{z_n}}{1+v_{z_n}}\\
& = \frac{\exp(r_{z_n})}{1+\exp(r_{z_n})}.
\end{align}
%so that $y_n \sim
%\operatorname{Bernoulli}(\operatorname{logistic}(r_{z_n}))$ as
%desired.  

%While we now perform inference of the augmented posterior
%$\{\gamma_l\}, \{W_n\}, L, \{c_l(\cdot)\}, \{\zeta_n\}|\{y_n\};\{x_n\}$, we of course also know, by
%construction, (up to a constant) the augmented posterior with the augmenting variables integrated out, namely
%the original posterior $\{\gamma_l\}, \{c_l(\cdot)\}|\{y_n\};\{x_n\}$.  This is useful
%if we wish to use collapsed Gibbs or Metropolis-Hastings steps during inference.

\subsubsection{Schedule of Updates}
Given the augmented model, we cycle through the following steps in the following deterministic order.  These will each be discussed in detail shortly.  Regarding notation, we will use use $\theta_{\operatorname{aug}}$ to refer to the parameters of the augmented model: $(L, \{c_l(\cdot)\}_{l=0}^{L-1}, K, \{\gamma_l\}_{l=0}^{L-1}, \{U_n\}_{n=1}^N, \{\zeta_n\}_{n=1}^N )$, so that Gibbs updates can be described more succinctly.

\textbf{Step 1} (Gibbs steps for each $\gamma_l$):\\
Sample $\grave{\gamma}_l \sim p_{\gamma_l}(\cdot|(\theta_{\operatorname{aug}}\setminus \gamma_l),\{y_n\}_{n=1}^N;\{x_n\}_{n=1}^N)$ \\for $l=0,\dots,L-1$.

\textbf{Step 2} (Gibbs step for $K$):\\
Sample $\grave{K} \sim p_K(\cdot|(\theta_{\operatorname{aug}}\setminus K),\{y_n\}_{n=1}^N;\{x_n\}_{n=1}^N)$.

\textbf{Step 3} (Collapsed Metropolis Hastings Step):\\
Perform Metropolis-Hastings update over $(L,\{c_l(\cdot)\}_{l=0}^{L-1})$, under the original model from Equation (\ref{eqn:posterior}). This can be viewed as a collapsed Metropolis-Hastings step, where the collapsed parameters are the augmenting variables $\{U_n\}_{n=1}^{N},\{\zeta_n\}_{z=1}^{N}$. 

\textbf{Step 4} (Gibbs step for $(\{\zeta_n\}_{n=1}^{N}, \{U_n\}_{n=1}^{N})$):\\
\emph{Jointly} sample \begin{eqnarray*}
\lefteqn{(\{\grave{\zeta}_n\}_{n=1}^{N}, \{\grave{U}_n\}_{n=1}^{N})}
\\&\sim p_{\{\zeta_n\}_n, \{U_n\}}(\cdot|\theta_{\operatorname{aug}}\setminus (\{\zeta_n\}_n,\{U_n\}_n), \{y_n\}_n;\{x_n\}_n).
%\intertext{(Gibbs step for $(\{\zeta_n\}_{n=1}^{N}, \{U_n\}_{n=1}^{N})$}.
\end{eqnarray*}

Mixing Gibbs and collapsed Metropolis-Hastings sampling steps requires special care to ensure the Markov chain is proper in that it possesses the desired stationary distribution.  We refer to \citet{van2011partially} for details, but do note that after the collapsed Metropolis-Hastings step, first performing a Gibbs update of $\{\zeta_n\}_{n=1}^{N}$, and then a separate Gibbs update for $\{U_n\}_{n=1}^{N}$ (or in reverse) would not have been proper.

\subsubsection{Update Details}

We now elaborate on each step of the update schedule:

%\noindent \emph{Step 1}\\
\paragraph{Step 1}
In this augmented model, the full conditional distribution of each $\gamma_l$
%,$p_{\gamma_l}(\cdot|\gamma_{0,\dots,l-1,l+1,\dots,L-1},\{U_n\}_{n=1}^N, L, \{c_l(\cdot)\}_{l=0}^{L-1},K; \{x_n\}_{n=1}^N)$ 
is Gamma
distributed, so that
it can be sampled from directly.  Let, for $l=0,\dots,L-1$
\begin{align}
\sigma_k^{(l)} =
\begin{cases}
K\Pi_{i=k,i \neq l}^{L-1} \gamma_i & \text{for} \enskip 0 \leq k \leq l\\
0 & \text{for} \enskip l < k \leq L.
\end{cases}  
\end{align}
Then, it can be derived that 
\begin{align}
\lefteqn{\gamma_l | (\theta_{\operatorname{aug}}\setminus \gamma_l), \{y_n\}_{n=1}^N;\{x_n\}_{n=1}^N}\\
&\sim \text{Gamma} \left(\alpha_l + \sum_{n=1}^N 1[z_n \leq l]
U_n, \beta_l + \sum_{n=1}^N \zeta_n\sigma_{z_n}^{(l)} \right),\nonumber
\end{align}
where $\alpha_l, \beta_l$ govern the prior of $\gamma_l$ and $z_n$ as described in Equation (\ref{eqn:z_n}) denotes the rule a datum belongs to.

%\hfill \break
%\noindent \emph{Step 2}\\
\paragraph{Step 2}
Similarily, let
\begin{align}
o_k =
\begin{cases}
\Pi_{i=k}^{L-1} \gamma_i & \text{for} \enskip 0 \leq k \leq L-1\\
1 & \text{for} \enskip k = L.
\end{cases}  
\end{align}
Then
\begin{align}
\lefteqn{K|(\theta_{\operatorname{aug}} \setminus K), \{y_n\}_{n=1}^N;\{x_n\}_{n=1}^N}\\
&\sim
 \text{Gamma} \left(\alpha_K + \sum_{n=1}^N U_n, \beta_K +\sum_{n=1}^N \zeta_no_{z_n} \right).
\end{align}

%\subsubsection{Sampling $W_n$}
%If $y_n=0$, then $W_n$ is constrained to be 0, due to equation
%(\ref{eqn:w}).  In the other case:

%\begin{align}
%W_n |\theta \setminus \{W_n\}, y_n=1; \{x_n\} \sim
%\operatorname{truncated\_poisson}(\zeta_n*v_{z_n}, 1)
%\end{align}

%where $\operatorname{truncated\_poisson(\lambda, K)}$ denotes a poisson
%distribution with rate $\lambda$ truncated at $K$.

%\subsubsection{Sampling $\zeta_n$}
%Once again due to gamma-poisson conjugacy, we have

%\begin{align}
%\zeta_n|\theta \setminus \{\zeta_n\}, \{y_n\}; \{x_n\}
%\sim \operatorname{gamma}(1+W_n, 1 + v_{z_n}) \label{eqn:zeta_gibbs}
%\end{align}

%\emph{Step 3}\\
\paragraph{Step 3}
The reason for using a collapsed rather than regular Metropolis-Hastings step was to improve chain mixing.  The Metropolis-Hastings proposal distributions over $(L$,$\{c_l(\cdot)\}_{l=0}^{L-1})$ are exactly as in the proposal distribution used to generate a successor state in the simulated annealing we used to find the MAP decision list.  The only difference is that if the $\operatorname{ADD}$ operation is chosen and a rule $c(\cdot)$ inserted at position $k$ in the rule list, then sample $\grave{\gamma}_{k} \sim \operatorname{Gamma}(\alpha_{k}, \beta_{k})$, the prior distribution of $\gamma_{k}$, and insert it at position $k$ in $\{\gamma_l\}_{l=0}^{L-1}$.  Thus, we simply provide the Metropolis-Hastings $Q$ probabilities.  For simplicity, we do so for the case when the weights $\{w_l\}_{l=0}^{|B|-1}$ associated with each $c(\cdot) \in B$ are equal.
\begin{eqnarray*}
Q(\{\grave{c}(\cdot)\}_{l=1}^{\grave{L}-1} ; \{c_l(\cdot)\}_{l=1}^{L-1})=
\begin{cases}
\frac{1}{(L+1)(|B|-L)}& \hspace*{-3mm} \textrm{if} \hspace*{.75mm} \operatorname{ADD}\\
\frac{1}{(|B|-L)L}&\hspace*{-3mm}\textrm{if} \hspace*{.75mm} \operatorname{REPLACE}\\
\frac{1}{L}&\hspace*{-3mm}\textrm{if}  \hspace*{.75mm}\operatorname{REMOVE}\\
\frac{2}{L(L-1)}&\hspace*{-3mm} \textrm{if} \hspace*{.75mm}\operatorname{SWAP}.
\end{cases}
\end{eqnarray*}

%\emph{Step 4}\\
\paragraph{Step 4}
In the full conditional distribution of $(\{\zeta_n\}_{n=1}^{N}, \{U_n\}_{n=1}^{N})$, the set of pairs of variables, $\{(U_n,\zeta_n)\}_{n=1}^{N}$ is mutually independent, due to conditioning.  Therefore to sample from it, it is sufficient to independently sample $(U_n,\zeta_n)$ for $n=1,\dots,N$.  It can be shown that the following sampling scheme samples from the full conditional distribution: For $n=1,\dots,N$, if $y_n=0$, set $\grave{U}_n = 0$ and sample 
\begin{align}
\grave{\zeta}_n &\sim \operatorname{exponential}(1+v_{z_n}).
%\end{align}
\shortintertext{Otherwise, sample}
%\begin{align}
\grave{U}_n &\sim 1 + \operatorname{Geometric}\left(\frac{1}{1+v_{z_n}}\right), \enskip \text{then}\\
\grave{\zeta}_n &\sim \operatorname{Gamma}\left(1+\grave{U}_n, 1+v_{z_n}\right).
\end{align}

\section{Simulation Studies}

We show that for simulated data generated by a known decision list, our simulated annealing procedure that searches for the MAP decision list, with high probability, recovers the true decision list.

Given observations with arbitrary features, and a collection of rules on those features, we can
construct a binary matrix where the rows represent observations and the columns represent rules, and the entry is 1 if the rule applies to that observation and 0 otherwise. We need
only simulate this binary matrix to represent the observations without losing generality. For
our simulations, we generated independent binary rule sets with 100 rules by setting each
feature value to 1 independently with probability 0.25.

\begin{figure}
\begin{centering}
\includegraphics[width=3in]{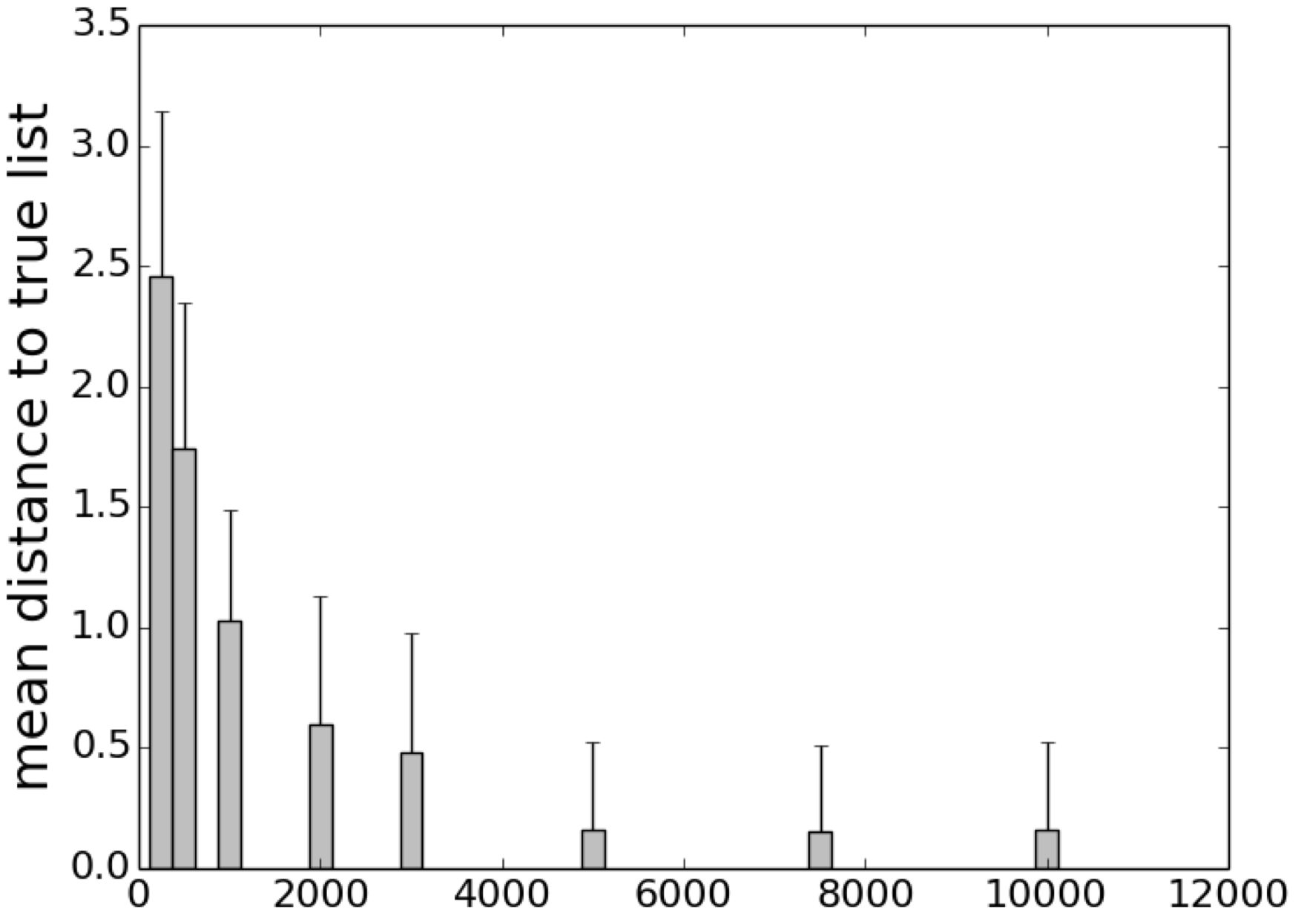}
\caption{Mean distance to true list decreases with sample size. \label{fig:sim_study}}
\end{centering}
\end{figure}

We generated a random decision list of size 5 by selecting 5 rules at random, and setting the $\gamma_0,\dots,\gamma_5$ so that the induced $p_0,\dots,p_5$ were roughly evenly spaced on the unit interval: $(.84, .70, .54, .40, .25, .14)$.  For each $N$, we performed the following procedure 100 times: generate the random rule matrix, random decision list, and assign it the aforementioned $\{\gamma_l\}$,  obtain an independent dataset of size $N$ by generating labels from this decision list, and then perform simulated annealing using the procedure described in Section \ref{sec:map} to obtain a point estimate of the decision list.  We then calculate the edit distance of the decision list returned by simulated annealing to the true decision list.  Figure \ref{fig:sim_study} shows the average distance over those 100 replicates, for each $N$.  We ran simulated annealing for $5000$ steps in each replicate, and used a gradual cooling schedule. %with slow temperature decrease.  %Further, we found that a cooling schedule with a slow temperature decrease balanced exploration and exploitation sufficiently to recover the true list. %and thus for both this simulation study and our experiments on real data, we used a constant temperature of 0 for simplicity.

\section{Experiments}
Our main experimental result is an application of Falling Rule Lists to predict 30 day hospital readmission from an ongoing collaboration with medical practitioners \citep[details to appear in][]{readmissionsData2014}. 
%We present the MAP model returned by simulated annealing to illustrate how one might make decisions using a falling rule list. 

Since we placed an extremely strong restriction on the characteristics of the predictive model (namely the monotonicity property, sparsity of rules, and sparsity of conditions per rule), we expect to lose predictive accuracy over unrestricted methods. The interpretability benefit may or not be sufficient to compensate for this, but this is heavily application-dependent. We have found several cases where there is no loss in performance (with a substantial gain in interpretability) by using a falling rule list instead of, say, a support vector machine, consistent with the observations of \cite{holte1993very} about very simple classifiers performing well. 

Later in this section, we aim to quantify the loss in predictive power from Falling Rule Lists over other methods by using an out-of-sample predictive performance evaluation. Specifically, we compare to several baseline methods on standard publicly available datasets to quantify the possible loss in predictive performance.

\subsection{Predicting Hospital Readmissions}

\begin{table*}[]
\begin{tabular}{lllll}
 & Conditions                            &                 & Probability &    Support  \\\hline
IF         & BedSores AND Noshow         & THEN readmissions risk is: & 33.25\% & 770  \\
ELSE IF    & PoorPrognosis AND MaxCare   & THEN readmissions risk is: & 28.42\% & 278  \\
ELSE IF    & PoorCondition AND Noshow    & THEN readmissions risk is: & 24.63\% & 337  \\
ELSE IF    & BedSores                    & THEN readmissions risk is: & 19.81\% & 308  \\
ELSE IF    & NegativeIdeation AND Noshow & THEN readmissions risk is: & 18.21\% & 291  \\
ELSE IF    & MaxCare                     & THEN readmissions risk is: & 13.84\% & 477  \\
%ELSE IF    & PoorPrognosis               & THEN readmissions risk is: & 7.11\%  & 394  \\
%ELSE IF    & Noshow                      & THEN readmissions risk is: & 6.00\%  & 983  \\
%ELSE IF    & MoodProblems                & THEN readmissions risk is: & 3.82\%  & 1152 \\
%ELSE       &                             & Readmissions risk is:      & 0.71\%  & 2954
ELSE IF    & Noshow                      & THEN readmissions risk is: & 6.00\%  & 1127  \\
ELSE IF    & MoodProblems                & THEN readmissions risk is: & 4.45\%  & 1325 \\
ELSE       &                             & Readmissions risk is:      & 0.88\%  & 3031
\end{tabular}
\caption{Falling rule list for patients with no multiple readmissions history. \label{table:readmissions_model}
}
\end{table*}

We applied Falling Rule Lists to preliminary readmissions data being compiled through a collaboration with a major hospital in the U.S. \citep{readmissionsData2014}, where the goal is to predict whether a patient will be readmitted to the hospital with 30 days, using data prior to their release.  The dataset contains features and binary readmissions outcomes for approximately 8,000 patients who had no prior history of readmissions. The features are very detailed, and include aspects like ``impaired mental status," ``difficult behavior," ``chronic pain," ``feels unsafe" and over 30 other features that might be predictive of readmission. As we will see, luckily a physician may not be required to collect this amount of detailed information to assess whether a given patient is at high risk for readmission.
%10,000 patients.  Because we found that the strongest predictor of future readmissions is past history of multiple readmissions, we removed 
%the roughly 20\% of 
%patients with a readmissions history to model the relationship between features and readmissions risk for those more interesting cases.

For these experiments and the experiments in the next section, no parameters were tuned in Falling Rule Lists (FRL), and the global hyperparameters were chosen as follows. We mined rules with a support of at least 5\% and a cardinality of at most 2 conditions per rule.  We assumed in the prior that conditioned on $L$, each rule had an equal chance of being in the rule list.  We set the prior of $\{\gamma_l\}|L$ to have noninformative and independent distributions of $\operatorname{gamma}(1,0.1)$, and the prior expected length of the decision list, $\lambda$, to be 8.  We performed simulated annealing search for $5000$ steps with a constant temperature of $1$ for simplicity. 

We measured out-of-sample performance using the AUROC 
%(the area under the received operator characteristic curve)
 from 5-fold cross validation where the MAP decision list from training was used to predict on each test fold in turn. We compared with SVM (with Radial Basis Function kernels), $\ell_2$ regularized logistic regression (Ridge regression, denoted LogReg), CART, and random forests (denoted RF), implemented in Python using the scikit-learn package. For SVM and logistic regression, hyperparameters were tuned with grid search in nested cross validation.

\begin{table}[]
\centering
\begin{tabular}{c|c}
Method & Mean AUROC (STD)\\
\hline \hline
FRL & .80 (.02) \\
NF\_FRL & .75 (.02)\\
NF\_GRD & .75 (.02)\\
RF & .79 (.03) \\
SVM & .62 (.06) \\
Logreg & .82 (.02) \\
Cart & .52 (.01)
\end{tabular}
\caption{AUROC values for readmission data
\label{table:readmissionAUC}}
\end{table}

% Furthermore, as mentioned in the introduction, decision lists consisting of the rules from an inductive logic programming method are not expected to exhibit strong predictive performance.  To show this empirically, we also the present the performance of two methods: We use nFoil \citep{landwehr2005nfoil} with the default settings (max number of clauses set to 2) to obtain a set of explanatory rules.  We then order those rules into a decision list two different ways, to obtain the two comparison methods: 1. by the empirical risk of each rule (denoted nFoilGreedy), and 2. by using the set of rules as the pre-mined rule set that the falling rule list algorithm accepts as input (denoted nFoilFRL).  Note that the risk probabilities in rule lists returned by nFoilGreedy are not necessarily decreasing monotonically, and that not all of the nFoil rules are necessarily in the rule list returned by nFoilFRL, should omission of a rule increase the posterior.

As discussed, decision lists consisting of rules from an inductive logic programming method are not expected to exhibit strong performance. We tested nFoil \citep{landwehr2005nfoil} with the default settings (max number of clauses set to 2) to obtain a set of rules. These rules were ordered in two different ways, to form two additional comparison methods: 1. by the empirical risk of each rule (denoted NF\_GRD), and 2. by using the set of rules as the pre-mined rule set that FRL accepts as input (denoted NF\_FRL).  Note that the risk probabilities in rule lists returned by NF\_GRD are not necessarily decreasing monotonically, and that not all of the nFoil rules are necessarily in the rule list returned by NF\_FRL, since omission of a rule can increase the posterior.

%For random forests, the number of trees was set to 

The AUROC's for the different methods are in Table \ref{table:readmissionAUC}, indicating that there was no loss in accuracy for using Falling Rule Lists on this particular dataset. For all of the training folds, the decision lists had a length of either 6 or 7 -- all very sparse.%length 6 and for remaining fold, the decision list was of length 7 - all very sparse.  

\begin{figure}
\begin{centering}
\includegraphics[width=3in]{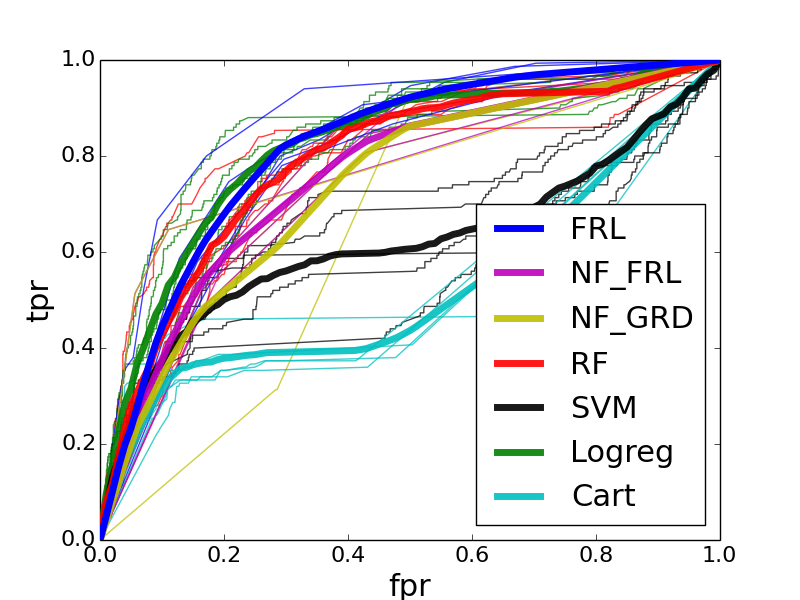}
\caption{ROC curves for readmissions prediction. \label{fig:readmissions_roc}}
\end{centering}
\end{figure}

Figure \ref{fig:readmissions_roc} shows ROC curves for all test folds for all methods. The mean ROC curves are bolded.  For this particular dataset, SVM RBF and CART did not perform well. It is unclear why SVM did not perform well, as cross-validation was performed for SVM; usually SVM's perform well when cross-validated (though it is definitely possible for them to have poor performance on some datasets -- on the other hand, CART often performs poorly relative to other methods, in our experience). As expected, the nFoil-based methods exhibited worse performance than our faIling rule list method.  FRL performed on par with the best method, despite its being limited to a very small number of features with the monotonic structure.
%python sklearn
% 66,777

Table \ref{table:readmissions_model} shows a point estimate obtained from training Falling Rule Lists on the full dataset, which took 88 seconds. The ``probability" column is the empirical probability of readmission for each rule; ``support" indicates the number of patients classified by that rule.

The model indicates that patients with bed sores and who have skipped appointments are the most likely to be readmitted. The other conditions used in the model include ``PoorPrognosis" meaning the patient is in need of palliative care services, ``PoorCondition"  meaning the patient exhibits frailty, signs of neglect or malnutrition, ``NegativeIdeation" which means suicidal or violent thoughts, and ``MaxCare" which means the patient needs maximum care (is non-ambulatory, confined to bed). This model lends itself naturally to decision-making, as one need only view the top of the list to obtain a characterization of high risk patients. 

\subsection{Performance on Public Datasets}

We performed an empirical comparison on several UCI datasets \citep{Bache+Lichman:2013}, using the above experimental setup. This allows us to quantify the loss in accuracy due to the restricted form of the model.

\begin{table}[b]
\centering
\begin{tabular}{c|c|c|c|c}
Method & Spam & Mamm & Breast & Cars\\
\hline \hline
FRL & .91(.01) & .82(.02)  &.95(.04) & .89(.08)\\
NF\_FRL & .90(.03) & .67(.03) & .70(.11) & .60(.21)\\
NF\_GRD & .91(.03) & .72(.04) & .82(.12) & .62(.20)\\
SVM & .97(.03) & .83(.01) &.99(.01) &.94(.08)\\
Logreg & .97(.03) & .85(.02) &.99(.01) & .92(.09)\\
CART & .88(.05) & .82(.02) &.93(.04) & .72(.17)\\
RF & .97(.03) & .83(.01) & .98(.01) & .92(.05)
\end{tabular}
\caption{AUROC value comparisons over datasets
\label{fig:uci_rocs}}
\end{table}

Table \ref{fig:uci_rocs} displays the results. As discussed earlier, we observed that even with the severe restrictions on the model, performance levels are still on par with those of other methods, and not often substantially worse. This is likely due to the benefits of not using a greedy splitting method, restricting to the space of mined rules, careful formulation, and optimization.

\begin{table}[]
\centering
\begin{tabular}{c|c|c|c|c|c}
Rule Miner & Radm & Spam & Mamm & Breast & Cars\\
\hline \hline
nFoil & 7 & 24 & 8 &5 &25\\
FPGrowth & 103& 550 & 43 & 147 & 190
\end{tabular}
\caption{Number of rules mined for each dataset.}
\label{num_rules}
\end{table}

Furthermore, FRL beats the nFoil based methods in performance on all the public datasets.  Again, the reason is that the set of rules from nFoil was found using a different objective, not intended to predict well when placed into a decision list.  Even with NF\_FRL, where any subset of the nFoil rules could be selected and placed in any order, performance was poor, showing the nFoil rule set did not contain rules that were useful for a falling rule list.  In fact, the set of nFoil rules was always much smaller than those from FPGrowth, overly restricting the hypothesis space.  We display in Table \ref{num_rules} the number of rules mined by FPGrowth and nFoil, on all datasets analyzed.

\begin{table}[]
\centering
\begin{tabular}{c|c|c|c}
Dataset & n & p & running time (sec)\\
\hline \hline
Radm & 7944 & 33 & 35\\
Spam & 4601 & 57 & 102\\
Mamm & 961 & 13 & 35\\
Breast & 683 & 26 & 42\\
Cars & 1728 & 21 & 39
\end{tabular}
\caption{Running time for 5000 simulated annealing steps, for each dataset.}
\label{running_times}
\end{table}

Finally, in Table \ref{running_times} we display for all datasets the running times to run the 5000 simulated annealing steps for MAP estimation.

%To construct full rule list on the mammographic mass dataset (n=961, p=13) took 35 seconds, for the spam dataset (n=4601, p=57) it took 102 seconds, for the breast cancer (n=683, p=27) dataset it took 35 seconds, and for the cars dataset (n=1728, p=21) it took 39 seconds. This method can be parallelized naturally by running multiple chains in parallel on separate processors.

%Figure \ref{fig:mammolist} from the introduction provides a rule list trained on the full mammographic mass dataset.

\section{Conclusion}
We present a new class of interpretable predictive models that could potentially have a major benefit in decision-making for some domains.
As nicely stated by the director of the U.S. National Institute of Justice \citep{pitfall}, an interpretable model that is actually used is better than one that is more accurate that sits on a shelf. We envision that models produced by FRL can be used, for instance, by physicians in third world countries who require models printed on laminated cards. In high stakes decisions (like medical decisions), it is important we know whether to trust the model we are using to make decisions; models like FRL help tell us when (and when not) to trust.

\textbf{Acknowledgment:} We gratefully acknowledge funding from Wistron, Siemens, and NSF-CAREER IIS-1053407.

%These methods can be used 
%by people making rapid decisions on the fly (like doctors). They could be used 
%by people who do not trust black box models, or by people who simply want to know reasons for predictions.

%Use unnumbered third level headings for the acknowledgements.  Allacknowledgements go at the end of the paper.  Be sure to omit any
%identifying information in the initial double-blind submission!

%\subsubsection*{References}

\bibliographystyle{authordate1}
\bibliography{bib1}
\thispagestyle{empty}

\end{document}